\documentclass[11pt,a4paper]{article}
\usepackage[hyperref]{emnlp-ijcnlp-2019}
\usepackage{times}
\usepackage{latexsym}
\usepackage{soul}
\usepackage{url}
\usepackage[utf8]{inputenc}
\usepackage{caption}
\usepackage{graphicx}
\usepackage{enumitem}
\usepackage{amsmath}

\usepackage{url}

\aclfinalcopy 

\title{A logical-based corpus for cross-lingual evaluation%
\thanks{~This study was financed in part by the Coordena\c{c}\~{a}o de Aperfei\c{c}oamento de Pessoal de N\'{i}vel Superior -- Brasil (CAPES) -- Finance Code 001; and Fapesp 2019/07665-4.}%
}

\author{
Felipe Salvatore$^1$,
Marcelo Finger$^1$\thanks{~Partly supported by Fapesp project 2014/12236-1 and CNPq grant PQ 303609/2018-4.}\and
R. Hirata Jr$^1$\thanks{~Partly supported by FAPESP projects 2015/01587-0, 2015/24485-9 and 2017/25835-9}\\
$^1$Department of Computer Science, Instituto de Matem\'atica e Estat\'istica, \\
University of S\~ao Paulo, Brazil\\
\{felsal, mfinger, hirata\}@ime.usp.br
}

\begin{document}
\maketitle
\begin{abstract}
At present, different deep learning models are presenting high accuracy on popular inference datasets such as SNLI, MNLI, and SciTail. However, there are different indicators that those datasets can be exploited by using some simple linguistic patterns. This fact poses difficulties to our understanding of the actual capacity of machine learning models to solve the complex task of textual inference. We propose a new set of syntactic tasks focused on contradiction detection that require specific capacities over linguistic logical forms such as: Boolean coordination, quantifiers, definite description, and counting operators. We evaluate two kinds of deep learning models that implicitly exploit language structure: recurrent models and the Transformer network BERT. We show that although BERT is clearly more efficient to generalize over most logical forms, there is space for improvement when dealing with counting operators. Since the syntactic tasks can be implemented in different languages, we show a successful case of cross-lingual transfer learning between English and Portuguese.
\end{abstract}

\section{Introduction}

Natural Language Inference (NLI) is a complex problem of Natural Language Understanding which is usually defined as follows: given a pair of textual inputs $P$ and $H$ we need to determine if $P$ \textit{entails} $H$, or $H$ \textit{contradicts} $P$, or $H$ and $P$ have no logical relationship (they are \textit{neutral}) \newcite{Fracas96}. $P$ and $H$, known as ``\textit{premise}" and ``\textit{hypothesis}" respectively, can be either simple sentences or full texts.

The task can focus either on the entailment or the contradiction part. The former, which is known as Recognizing Textual Entailment (RTE) \newcite{series/synthesis/2013Dagan}, classifies the pair $P$, $H$ in ``\textit{entailment}" or ``\textit{non-entailment}". The latter, which is know as Contradiction Detection (CD), classifies that pair in terms of ``\textit{contradiction}" or ``\textit{non-contradiction}". Independently of the form that we frame the problem, the concept of inference is the critical issue here.

With this formulation, NLI has been treated as a text classification problem suitable to be solved by a variety of machine learning techniques \newcite{Bowman15,Williams17}. Inference itself is also a complex problem. As shown in the following sentence pairs:

\begin{enumerate}
    \item ``\textit{A woman plays with my dog}", ``\textit{A person plays with my dog}"
    \item ``\textit{Jenny and Sally play with my dog}", ``\textit{Jenny plays with my dog}"  
\end{enumerate}

Both examples are cases of entailment, with different properties. In (1) the entailment is caused by the hypernym relationship between ``\textit{person}" and ``\textit{woman}". Example (2) deals with interpretation of the coordinating conjunction ``\textit{and}" as a Boolean connective. As (1) relies on the meaning of the noun phrases we call it  ``\textit{lexical inference}". As (2) is invariant under substitution we call it ``\textit{structural inference}". The latter is the focus of this work.

In this paper, we propose a new synthetic CD dataset that enables us to:
\begin{enumerate}
    \item compare the NLI accuracy of different deep learning models.
    \item diagnose the structural (logical and syntactic) competence of each model.
    \item verify the cross-lingual performance of each method.
\end{enumerate} 

The contributions presented in this paper are: i) the presentation of a structure oriented CD dataset; ii) the comparison of traditional neural recurrent models against the Transformer network BERT; iii) a success case of cross-lingual transfer learning for structural NLI between English and Portuguese.    

\section{Background and Related Work}

The size of NLI datasets has been increasing since the initial proposition of the FraCas test suit composed of $346$ examples \newcite{Fracas96}. Some old datasets like RTE-6 \newcite{Bentivogli2009} and SICK \newcite{Marelli2014}, with $16$K and $9.8$K examples, respectively, are relatively small if compared with the current ones like SNLI \newcite{Bowman15} and MNLI \newcite{Williams17}, with $570$K and $433$K examples, respectively. This increase was possible with the use of crowdsource platforms like the Amazon Mechanical Turk \newcite{Bowman15,Williams17}. The annotation performed by a formal semanticist, like in RTE 1-3 \newcite{Giampiccolo07}, was replaced with the generation of sentence pairs done by average English speakers. This change in dataset construction has been criticised with the argument that it is hard for an average speaker to produce different and creative examples of entailment and contradiction pairs \newcite{Gururangan18}. By looking at the hypothesis alone a simple text classifier can achieve an accuracy significantly better than a random classifier in datasets such as SNLI and MNLI. This was explained by a high correlation of occurrences of negative words (``\textit{no}", ``\textit{nobody}", ``\textit{never}", ``\textit{nothing}") in contradiction instances, and high correlation of generic words (such as ``\textit{animal}", ``\textit{instrument}", ``\textit{outdoors}") with entailment instances. Thus, despite of the large size of the corpora the task was easier to perform than expected \newcite{Poliak18}.

The new wave of pre-trained models \newcite{Howard18,Devlin18,Liu19} poses both a challenge and an opportunity for the NLI field. The large-scale datasets are close to being solved (the benchmark for SNLI, MNLI, and SciTail is $91.1\%$, $85.3\%/85.0\%$, and $94.1\%$, respectively, as reported in \newcite{Liu19}), giving the impression that NLI will become a trivial problem. The opportunity lies in the fact that, by using pre-trained models, training will no longer need such large datasets. Then we can focus our efforts in creating small, well-thought datasets that reflect the variety of inferential tasks, and so determine the real competence of a model. 

Here we present a collection of small datasets designed to measure the competence of detecting contradictions in structural inferences. 
We have chosen the CD task because it is harder for an average annotator to create examples of contradictions without excessively relying on the same patterns. At the same time, CD has practical importance since it can be used to improve consistency in real case applications, such as chat-bots \newcite{Welleck18}.

We choose to focus on structural inference because we have detected that the current datasets are not appropriately addressing this particular feature. In an experiment, we verify the deficiency reported in \newcite{Gururangan18,Glockner18}. First, we transformed the SNLI and MNLI datasets to a CD task. The transformation is done by converting all instances of entailment and neutral into non-contradiction, and by balancing the classes in both training and test data. Second, we applied a simple Bag-of-Words classifier, destroying any structural information. The accuracy was significantly higher than the random classifier, $63.9\%$ and $61.9\%$ for SNLI and MNLI, respectively. Even the recent dataset focusing on contradiction, Dialog NLI \newcite{Welleck18}, presents a similar pattern. The same Bag-of-Words model achieved $76.2\%$ accuracy in this corpus.

 Our approach of isolating structural forms by using synthetic data to analyze the logical and syntactical competence of different neural models is similar to \newcite{Bowman15b,Evans18,Tran18}. One main difference between their approach and ours is that we are interested in using a formal language as a tool for performing a cross-lingual analysis.

\section{Data Collection}

The different datasets that we propose are divided by tasks, such that each task 
introduces a new linguistic construct. Each task is designed by applying structurally dependent rules to automatically generate the sentence pairs. We first define the pairs in a formal language and then we use it to generate instances in natural language.  In this paper, we have decided to work with English and Portuguese.

There are two main reasons to use a formal language as a basis for the dataset. First, this approach allows us to minimize the influence of common knowledge and lexical knowledge, highlighting structural features. Second, we can obtain a structural symmetry between the English and Portuguese corpora.

Hence, our dataset is a tool to measure inference in two dimensions: one defined by the structural forms, which corresponds to different levels in our hierarchical corpus; and other defined by the instantiation of these forms in multiple natural languages.

\subsection{Template Language}

The \textit{template language} is a formal language used to generate instances of contradictions and non-contradictions in a natural language. 
This language is composed of two basic entities: people, $Pe = \{ x_1, x_2, ..., x_n\}$ and places, $Pl = \{p_1, p_2, ..., p_m\}$. We also define three binary relations: $V(x,y)$ , $x > y$, $x\geq y$. It is a simplistic universe with the intended meaning for binary relations such as ``$x$ \textit{has visited} $y$", ``$x$ \textit{is taller than} $y$" and ``$x$ \textit{is as tall as} $y$", respectively.

A \textit{realisation} of the template language $r$ is a function mapping $Pe$ and $Pl$ to nouns such that $r(Pe) \, \cap \, r(Pl) = \emptyset$; it also maps the relation symbols and logic operators to corresponding forms in some natural language.

Each task is defined by the introduction of a new structural and logical operator. We define the tasks in a hierarchical fashion: if a logical operator appears on 
a
task $n$, it can appear in any task $k$ (with $k>n$). The main advantage of our approach compared to other datasets is that we can isolate the occurrences of each operator to have a clear notion in what forces the models to fail (or succeed).

For each task, we provide training and test data with 10K and 1K examples, respectively. All data is balanced; and, as usual, the model's accuracy is evaluated on the test data. To test the model's generalization capability, we have defined two distinct realization functions $r_{train}$ and $r_{test}$ such that $r_{train}(Pe) \, \cap \,  r_{test}(Pe) = \emptyset$ and $r_{train}(Pl) \, \cap \,  r_{test}(Pl)  = \emptyset$. For example, in the English version $r_{train}(Pe)$ and $r_{train}(Pl) $ are composed of common English masculine names and names of countries, respectively. Similarly, $r_{test}(Pe)$ and $r_{test}(Pl)$ are composed of feminine names and names of cities from the United States. In the Portuguese version we have done a similar construction, using common masculine and feminine names together with names of countries and names of Brazilian cities.

\subsection{Data Generation}

A logical rule can be seen as a mapping that transforms a premise $P$ into a conclusion $C$. To obtain examples of contradiction we start with a premise $P$ and define $H$ as the negation of $C$. The examples of non-contradiction are different negations that do not necessarily violate $P$.  We repeat this process for each task. What defines the difference from one task to another is the introduction of logical and linguist operators, and subsequently, new rules.
We have used more than one template pair to define each task; however, for the sake of brevity, in the description below we will give only a brief overview of each task.

The full dataset in both languages, together with the code to generate it and the detailed list of all templates, can be found online \newcite{Salvatore19}.

\textbf{Task 1: Simple Negation}  We introduce the negation operator $\lnot$, ``\textit{not}". The premise $P$ is a collection of facts about some agents visiting different places. Example, $P:=\{V(x_1, p_1), V(x_2, p_2)\}$ (``\textit{Charles has visited Chile, Joe has visited Japan}"). The hypothesis $H$ can be either a negation of one fact that appears in $P$, $\lnot V(x_2, p_2)$ (``\textit{Joe didn't visit Japan}"); or a new fact not related to $P$, $\lnot V(x, p)$ (``\textit{Lana didn't visit France}"). The number of facts that appear in $P$ vary from two to twelve.

\textbf{Task 2: Boolean Coordination} In this task, we add the Boolean conjunction $\land$, the coordinating conjunction ``\textit{and}". Example, $P: = \{ V(x1,p)\land V(x2,p)\land V(x3,p) \}$ (``\textit{Felix, Ronnie, and Tyler have visited Bolivia}"). The new information $H$ can state that one of the mentioned agents did not travel to a mentioned place, $\lnot V(x_3, p)$ (``\textit{Tyler didn't visit Bolivia}"). Or it can represent a new fact, $\lnot V(x, p)$ (``\textit{Bruce didn't visit Bolivia}").
 
\textbf{Task 3: Quantification} By adding the quantifiers $\forall$ and $\exists$, ``\textit{for every}" and ``\textit{some}", respectively, we can construct example of inferences that explicitly exploit the difference between the two basic entities, people and places. Example, $P$ states a general fact about all people, $P:= \{ \forall x \forall p V(x,p)\}$ (``\textit{Everyone has visited every place}") . $H$ can be the negation of one particular instance of $P$, $ \lnot V(x,p)$ (``\textit{Timothy didn't visit El Salvador}"). Or a fact that does not violate $P$, $ \lnot V(x,x_1)$ (``\textit{Timothy didn't visit Anthony}"). 

\textbf{Task 4: Definite Description} One way to test if a model can capture reference is by using definite description, i.e., by adding the operator $\iota$  to perform description and the equality relation $=$. Hence, $x = \iota y Q(y)$ is to be read as ``\textit{$x$ is the one that has property Q}". Here we describe one property of one agent and ask the model to combine the description with a new fact. For example, $P := \{ x_1 = \iota y \forall p V(y, p) , V(x_1, x_2)\}$ (``\textit{Carlos is the person that has visited every place, Carlos has visited John}"). Two new hypotheses can be introduced: $\lnot V(x_1, p)$  (``\textit{Carlos did not visit Germany}") or $\lnot V(x_2, p)$  (``\textit{John did not visit Germany}"). Only the first hypothesis is a contradiction. Although the names ``\textit{Carlos}" and ``\textit{John}" appear on the premise, we expected the model to relate the property ``\textit{being the one that has visited every place}" to ``\textit{Carlos}" and not to ``\textit{John}".
 
 \textbf{Task 5: Comparatives} In this task we are interested to know if the model can recognise a basic property of a binary relation: transitivity. The premise is composed of a collection of simple facts $P:= \{ x_1 > x_2, x_2 > x_3\}$. (``\textit{Francis is taller than Joe, Joe is taller than Ryan}"). Assuming the transitivity of $>$, the hypothesis can be a consequence of $P$, $x_1 > x_3$ (``\textit{Francis is taller than Ryan}"), or a fact that violates the transitivity property, $x_3 > x_1$ (``\textit{Ryan is taller than Francis}"). The size of the $P$ varies from four to ten. Negation is not employed here.
 
 \textbf{Task 6: Counting} In Task 3 we have added only the basic quantifiers $\forall$ and $\exists$, but there is a broader family of operators called \textit{generalised quantifiers}. In this task we introduce the counting quantifier $\exists_{=n}$ (``\textit{exactly} $n$"). Example, $P:= \{ \exists_{=3} p V(x_1, p) \land \exists_{=2} x V(x_1, x) \}$ (``\textit{Philip has visited only three places and only two people}"). $H$ can be an information consistent with $P$, $V(x_1, x_2)$ (``\textit{Philip has visited John}"), or something that contradicts $P$, $V(x_1, x_2)\land V(x_1, x_3)\land V(x_1, x_4)$ (``\textit{Philip has visited John, Carla, and Bruce}"). We have added counting quantifiers corresponding to numbers from one to thirty.

 \textbf{Task 7: Mixed} In order to guarantee variability, we created a dataset composed of different samples of the previous tasks.
 
 Basic statistics for the English and Portuguese realisations of all tasks can be found in Table \ref{task_describe}. 
 
 \begin{table}[h!]
\begin{tabular}{|l|l|l|l|l|llllllll}
\cline{1-5}
Task        & \begin{tabular}[c]{@{}l@{}}Vocab\\ size\end{tabular} & \begin{tabular}[c]{@{}l@{}}Vocab\\ inter-\\section\end{tabular} & \begin{tabular}[c]{@{}l@{}}Mean\\ input\\ length\end{tabular} & \begin{tabular}[c]{@{}l@{}}Max\\ input\\ length\end{tabular} &  &  & \multicolumn{1}{c}{} &  &  &  &  &  \\ \cline{1-5}
1 (Eng)& 3561                                                      & 77                                                                & 230.6                                                         & 459                                                          &  &  &                      &  &  &  &  &  \\ \cline{1-1}
2 (Eng) & 4117                                                      & 128                                                               & 151.4                                                         & 343                                                          &  &  & \multicolumn{1}{c}{} &  &  &  &  &  \\ \cline{1-1}
3 (Eng) & 3117                                                      & 70                                                                & 101.5                                                         & 329                                                          &  &  & \multicolumn{1}{c}{} &  &  &  &  &  \\ \cline{1-1}
4 (Eng) & 1878                                                      & 62                                                                & 100.81                                                        & 134                                                          &  &  & \multicolumn{1}{c}{} &  &  &  &  &  \\ \cline{1-1}
5 (Eng) & 1311                                                      & 25                                                                & 208.8                                                         & 377                                                          &  &  &                      &  &  &  &  &  \\ \cline{1-1}
6 (Eng) & 3900                                                      & 150                                                               & 168.4                                                         & 468                                                          &  &  &                      &  &  &  &  &  \\ \cline{1-1}
7 (Eng) & 3775                                                      & 162                                                               & 160.6                                                         & 466                                                          &  &  &                      &  &  &  &  &  \\ \cline{1-1}
1 (Pt)  & 7762                                                      & 254                                                               & 209.4                                                         & 445                                                          &  &  &                      &  &  &  &  &  \\ \cline{1-1}
2 (Pt)  & 9990                                                      & 393                                                               & 148.5                                                         & 388                                                          &  &  &                      &  &  &  &  &  \\ \cline{1-1}
3 (Pt)  & 5930                                                      & 212                                                               & 102.7                                                         & 395                                                          &  &  &                      &  &  &  &  &  \\ \cline{1-1}
4 (Pt)  & 5540                                                      & 135                                                               & 91.8                                                          & 140                                                          &  &  &                      &  &  &  &  &  \\ \cline{1-1}
5 (Pt)  & 5970                                                      & 114                                                               & 235.2                                                         & 462                                                          &  &  &                      &  &  &  &  &  \\ \cline{1-1}
6 (Pt)  & 9535                                                      & 386                                                               & 87.8                                                          & 531                                                          &  &  &                      &  &  &  &  &  \\ \cline{1-1}
7 (Pt)  & 8880                                                      & 391                                                               & 159.9                                                         & 487                                                          &  &  &                      &  &  &  &  &  \\ \cline{1-5}
\end{tabular}
\caption{\label{task_describe} Task description. Column 1 presents two realizations of the described tasks - one in English (Eng) and the other in Portuguese (Pt). Column 2 presents the vocabulary size for the task. Column 3 presents the number of words that occurs both in the training and test data. Column 4 presents the average length in words of the input text (the concatenation of $P$ and $H$). Column 5 presents the maximum length of the input text.}
\end{table}

Since we are using a large number of facts in $P$, the input text is longer than the ones presented in average NLI datasets.

\section{Models and Evaluation}

To evaluate the accuracy of each CD task we employed three kinds of models:

\textbf{Baseline} The baseline model (Base) is a Random Forest classifier that models the input text, the concatenation of $P$ and $H$, using the Bag-of-Words representation. Since we have constructed the dataset centered on the notion of structure-based contradictions, we believe that it should perform slightly better than random. At the same time, by using such baseline, we can certify if the proposed tasks are indeed requiring structural knowledge.  

\textbf{Recurrent Models} The dominant family of neural models in Natural Language Processing specialised in modelling sequential data is the one composed by the \textit{Recurrent Neural Networks} (RNNs) and its variations, \textit{Long Short-Term Memory} (LSTM), and \textit{Gated Recurrent Unit} (GRU) \newcite{Goldberg15}. We consider both the standard and the bidirectional variants of this family of models. As input for these models, we use the concatenation of $P$ and $H$ as a single sentence. 

Traditional multilayer recurrent models are not the best choice to improve the benchmark on NLI \newcite{Glockner18}. However, in recent works, it has been reported that recurrent models achieve a better performance than Transformer-based models to capture structural patterns for logical inference \newcite{Evans18,Tran18}. We want to investigate if the same result can be achieved using our tasks as the base of comparison.

\textbf{Transformer-based Models}
A recent non-recurrent family of neural models known as \textit{Transformer networks} was introduced in \newcite{Vaswani17}. Different from the recurrent models that recursively summarizes all previous input into a single representation, the Transformer network employes a self-attention mechanism to directly attend to all previous inputs (more details of this architecture can be found in \newcite{Vaswani17}). Although, by performing regular training using this architecture alone we do not see surprising results in inference prediction  \newcite{Evans18,Tran18}, when we pre-trained a Transformer network in the language modeling task and fine-tuned afterwards on an inference task we see a significant improvement \newcite{Devlin18}.

Among the different Transformer-based models we will focus our analysis on the multilayer bidirectional architecture known as \textit{Bidirectional Encoder Representation from Transformers} (BERT) \newcite{Devlin18}. This bidirectional model,  pre-trained as a masked language model and as a next sentence predictor, has two versions: BERT${_\textsc{base}}$ and BERT$_\textsc{large}$. The difference lies in the size of each architecture, the number of layers and self-attention heads. Since BERT$_\textsc{large}$ is unstable on small datasets \newcite{Devlin18} we have used only BERT${_\textsc{base}}$.

The strategy to perform NLI classification using BERT is the same the one presented in \newcite{Devlin18}: together with the pair $P,H$ we add new special tokens [CLS] (classification token) and [SEP] (sentence separator). Hence, the textual input is the result of the concatenation: [CLS] $P$ [SEP] $H$ [SEP]. After we obtain the vector representation of the [CLS] token, we pass it through a classification layer to obtain the prediction class (contradiction / non-contradiction). We fine-tune the model for the CD task in a standard way, the original weights are co-trained with the weights from the new layer.

By comparing BERT with other models we are not only comparing different architectures but different techniques of training. The baseline model uses no additional information. The recurrent models use only a soft version of transfer learning with fine-tuning of pre-trained embeddings (the fine-tuning of one layer only). On the other side, BERT is pre-trained on a large corpus as a language model. It is expected that this pre-training helps the model to capture some general properties of language \newcite{Howard18}. Since the tasks that we proposed are basic and cover very specific aspects of reasoning, we can use it to evaluate which properties are being learned in the pre-training phase.

The simplicity of the tasks motivated us to use transfer-learning differently: instead of simply using the multilingual version of BERT\footnote{Multilingual BERT is a model trained on the concatenation of the entire Wikipedia from 100 languages, Portuguese included. \url{https://github.com/google-research/bert/blob/master/multilingual.md}} and fine-tune it on the Portuguese version of the tasks, \textit{we have decided to check the possibility of transferring structural knowledge from  high-resource languages (English / Chinese) to Portuguese.}

This can be done because for each pre-trained model there is a tokenizer that transforms the Portuguese input into a collection of tokens that the model can process. Thus, we have decided to use the regular version of BERT trained on an English corpus (BERT$_{eng}$), the already mentioned Multilingual BERT (BERT$_{mult}$), and the version of the BERT model trained on a Chinese corpus (BERT$_{chi}$).

We hypothesize that \textit{most structural patterns learned by the model in English can be transferred to Portuguese}. By the same reasoning, we believe that BERT$_{chi}$ should perform poorly. Not only the tokenizer associated to BERT$_{chi}$ will add noise to the input text, but also Portuguese and Chinese are grammatically different; for example, the latter is overwhelmingly right-branching while the former is more mixed \newcite{Levy03}.

\subsection{Experimental settings}

Given the above considerations, four research questions arose:

\begin{enumerate}[label=(\roman*)]
    \item\textit{How the different models perform on the proposed tasks?}
    
    \item \textit{How much each model rely on the occurrence of non-logical words?}

    \item \textit{Can cross-lingual transfer learning be successfully used for the Portuguese realization of those tasks?}
    
    \item \textit{Is the dataset biased? Are the models learning some unexpected text pattern?}
\end{enumerate}

To answer those questions, we evaluated the models performance in four different ways:

\begin{enumerate}[label=(\roman*)]
    \item Each model was trained on different proportions of the dataset. In this case, $r_{train}(Pe) \cap r_{test}(Pe)= \emptyset$ and  $r_{train}(Pl) \cap r_{test}(Pl)= \emptyset$.
    \item We have trained the models on a version of the dataset where we allow full intersection of the train and test vocabulary, i.e., $r_{train}(Pe) = r_{test}(Pe)$ and  $r_{train}(Pl) = r_{test}(Pl)$.
    \item For the Portuguese corpus, we have fine-tuned the three pre-trained models mentioned previously: BERT$_{eng}$, BERT$_{mult}$, and BERT$_{chi}$.
    \item We have trained the best model from (i) on the following modified versions of the dataset:
    \begin{enumerate}[label=(\alph*)]
        \item \textit{Noise label} - each pair $P$, $H$ is unchanged but we randomly labeled the pair as contradiction or non-contradiction.
        \item \textit{Premise only} - we keep the labels the same and omit the hypothesis $H$.
        \item  \textit{Hypothesis only} - the premise $P$ is removed, but the labels remain intact. 
    \end{enumerate}
\end{enumerate}

\subsection{Implementation}

All deep learning architectures were implemented using the Pytorch library \newcite{paszke17}. To make use of the pre-trained version of BERT we have based our implementation on the public repository \url{https://github.com/huggingface/pytorch-pretrained-BERT}. 

The different recurrent architectures were optimized with Adam \newcite{Kingma14}. We have used pre-trained word embedding from Glove \newcite{Pennington14} and Fasttext \newcite{Joulin16}, but we also used random initialized embeddings. We random searched across embedding dimensions in $[10, 500]$, hidden layer size of the recurrent model in $[10, 500]$, number of recurrent layer in $[1, 6]$, learning rate in $[0,1]$, dropout in $[0,1]$ and batch sizes in $[32,128]$.

The hyperparameter search for BERT follows the one presented in \newcite{Devlin18} that uses Adam with learning rate warmup and linear decay.

We randomly searched the learning rate in $[2 \cdot 10^{-5}, 5 \cdot 10^{-5}]$, batch sizes in $[16,32]$ and number of epochs in $[3,4]$. 

All the code for the experiments is public available \newcite{Salvatore19}.

\subsection{Results}

\textit{How the different models perform on the proposed tasks?}

In most of the tasks, BERT$_{eng}$ presents a clear advantage when compared to all other models. Tasks 3 and 6 are the only ones where the difference in accuracy between BERT$_{eng}$ and the recurrent models is small, as can be seen in Table \ref{Tab:Class}. Even when we look at BERT$_{eng}$'s results on the Portuguese corpus, which are slightly worse when compared to the English one, we still see a similar pattern. 

\begin{figure}[h!]
    \centering
    \includegraphics[width=0.49\textwidth]{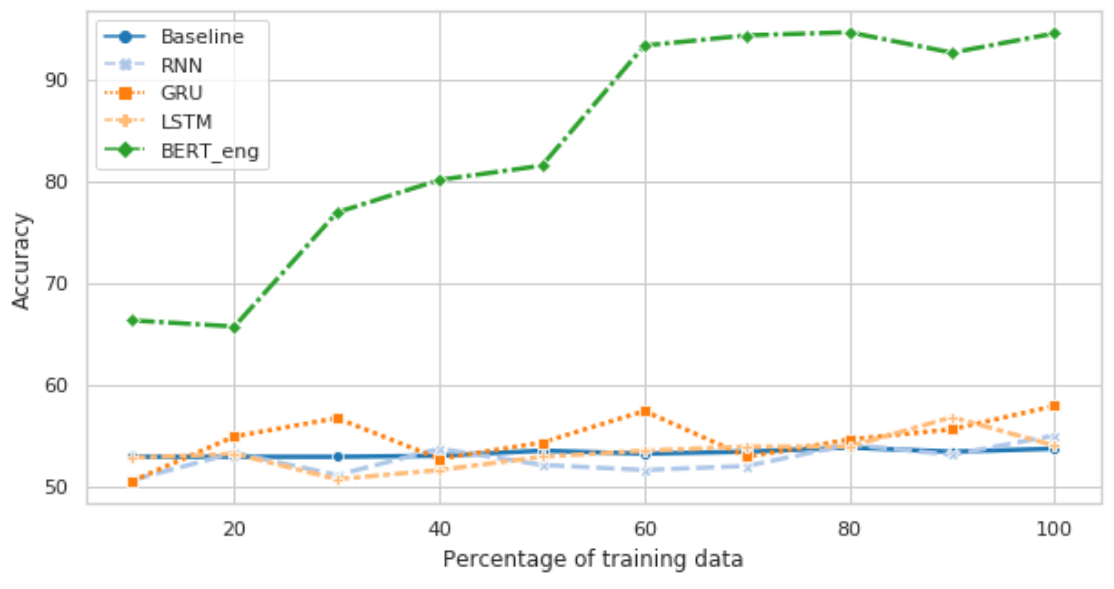}
  \caption{\label{Fig:acc1} Results of the experiment (i), accuracy for each model on different data proportions (English corpus)}
\end{figure}

Figure \ref{Fig:acc1} shows that BERT$_{eng}$ is the only model improved by training on more data. All other models remain close to random independently of the amount of training data.

\begin{figure}[h!]
    \centering
    \includegraphics[width=0.49\textwidth]{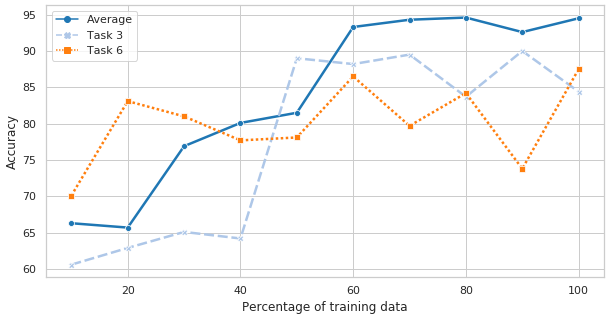}
  \caption{\label{Fig:acc2} Results of the experiment (i), BERT$_{eng}$'s accuracy on the different different tasks (English corpus)}
\end{figure}

Accuracy improvement over training size indicates the difference in difficulty of each task. On the one hand, Tasks 1, 2 and 4 are practically solved by BERT using only 4K examples of training ($99.5\%$, $99.7\%$, $97.6\%$ accuracy, respectively). On the other hand, the results for Tasks 3 and 6 remain below average, as seen in Figure \ref{Fig:acc2}.

\begin{table}[h!]
\begin{tabular}{llllllllllllll}
\multicolumn{1}{c}{}                &                               &                           &                           &                           &                           &  &  &                      &  &  &  &  &  \\
                                    &                               &                           &                           &                           &                           &  &  &                      &  &  &  &  &  \\ \cline{1-6}
\multicolumn{1}{|l|}{Task}              & \multicolumn{1}{l|}{Base} & \multicolumn{1}{l|}{RNN}  & \multicolumn{1}{l|}{GRU}  & \multicolumn{1}{l|}{LSTM} & \multicolumn{1}{l|}{BERT} &  &  & \multicolumn{1}{c}{} &  &  &  &  &  \\ \cline{1-6}
\multicolumn{1}{|l|}{1 (Eng)}  & \multicolumn{1}{l|}{52.1}     & \multicolumn{1}{l|}{50.1} & \multicolumn{1}{l|}{50.6} & \multicolumn{1}{l|}{50.4} & \multicolumn{1}{l|}{\textbf{99.8}} &  &  &                      &  &  &  &  &  \\ \cline{1-1}
\multicolumn{1}{|l|}{2 (Eng)}  & \multicolumn{1}{l|}{50.7}     & \multicolumn{1}{l|}{50.2} & \multicolumn{1}{l|}{50.2} & \multicolumn{1}{l|}{50.8} & \multicolumn{1}{l|}{\textbf{100}}  &  &  & \multicolumn{1}{c}{} &  &  &  &  &  \\ \cline{1-1}
\multicolumn{1}{|l|}{3 (Eng)}  & \multicolumn{1}{l|}{63.5}     & \multicolumn{1}{l|}{50.3} & \multicolumn{1}{l|}{66.1} & \multicolumn{1}{l|}{63.5} & \multicolumn{1}{l|}{\textbf{90.5}} &  &  & \multicolumn{1}{c}{} &  &  &  &  &  \\ \cline{1-1}
\multicolumn{1}{|l|}{4 (Eng)}  & \multicolumn{1}{l|}{51.0}     & \multicolumn{1}{l|}{51.7} & \multicolumn{1}{l|}{52.7} & \multicolumn{1}{l|}{51.6} & \multicolumn{1}{l|}{\textbf{100}}  &  &  & \multicolumn{1}{c}{} &  &  &  &  &  \\ \cline{1-1}
\multicolumn{1}{|l|}{5 (Eng)}  & \multicolumn{1}{l|}{50.6}     & \multicolumn{1}{l|}{50.1} & \multicolumn{1}{l|}{50.2} & \multicolumn{1}{l|}{50.2} & \multicolumn{1}{l|}{\textbf{100}}  &  &  &                      &  &  &  &  &  \\ \cline{1-1}
\multicolumn{1}{|l|}{6 (Eng)}  & \multicolumn{1}{l|}{55.5}     & \multicolumn{1}{l|}{84.4} & \multicolumn{1}{l|}{82.7} & \multicolumn{1}{l|}{75.1} & \multicolumn{1}{l|}{\textbf{87.5}} &  &  &                      &  &  &  &  &  \\ \cline{1-1}
\multicolumn{1}{|l|}{7 (Eng)}  & \multicolumn{1}{l|}{54.1}     & \multicolumn{1}{l|}{50.9} & \multicolumn{1}{l|}{53.7} & \multicolumn{1}{l|}{50.0} & \multicolumn{1}{l|}{\textbf{94.6}} &  &  &                      &  &  &  &  &  \\ \cline{1-6}
\multicolumn{1}{|l|}{Avg.} & \multicolumn{1}{l|}{53.9}     & \multicolumn{1}{l|}{55.4} & \multicolumn{1}{l|}{58.0} & \multicolumn{1}{l|}{56.2} & \multicolumn{1}{l|}{\textbf{96.1}} &  &  &                      &  &  &  &  &  \\ \cline{1-6}
\multicolumn{1}{|l|}{1 (Pt)}   & \multicolumn{1}{l|}{53.9}     & \multicolumn{1}{l|}{50.1} & \multicolumn{1}{l|}{50.2} & \multicolumn{1}{l|}{50.0} & \multicolumn{1}{l|}{\textbf{99.9}} &  &  &                      &  &  &  &  &  \\ \cline{1-1}
\multicolumn{1}{|l|}{2 (Pt)}   & \multicolumn{1}{l|}{49.8}     & \multicolumn{1}{l|}{50.0} & \multicolumn{1}{l|}{50.0} & \multicolumn{1}{l|}{50.0} & \multicolumn{1}{l|}{\textbf{99.9}} &  &  & \multicolumn{1}{c}{} &  &  &  &  &  \\ \cline{1-1}
\multicolumn{1}{|l|}{3 (Pt)}   & \multicolumn{1}{l|}{61.7}     & \multicolumn{1}{l|}{50.0} & \multicolumn{1}{l|}{70.6} & \multicolumn{1}{l|}{50.1} & \multicolumn{1}{l|}{\textbf{78.7}} &  &  & \multicolumn{1}{c}{} &  &  &  &  &  \\ \cline{1-1}
\multicolumn{1}{|l|}{4 (Pt)}   & \multicolumn{1}{l|}{50.9}     & \multicolumn{1}{l|}{50.0} & \multicolumn{1}{l|}{50.4} & \multicolumn{1}{l|}{50.0} & \multicolumn{1}{l|}{\textbf{100}}  &  &  & \multicolumn{1}{c}{} &  &  &  &  &  \\ \cline{1-1}
\multicolumn{1}{|l|}{5 (Pt)}   & \multicolumn{1}{l|}{49.9}     & \multicolumn{1}{l|}{50.1} & \multicolumn{1}{l|}{50.8} & \multicolumn{1}{l|}{50.0} & \multicolumn{1}{l|}{\textbf{99.8}} &  &  &                      &  &  &  &  &  \\ \cline{1-1}
\multicolumn{1}{|l|}{6 (Pt)}   & \multicolumn{1}{l|}{58.9}     & \multicolumn{1}{l|}{66.4} & \multicolumn{1}{l|}{\textbf{79.7}} & \multicolumn{1}{l|}{67.2} & \multicolumn{1}{l|}{79.1} &  &  &                      &  &  &  &  &  \\ \cline{1-1}
\multicolumn{1}{|l|}{7 (Pt)}   & \multicolumn{1}{l|}{55.4}     & \multicolumn{1}{l|}{51.1} & \multicolumn{1}{l|}{51.6} & \multicolumn{1}{l|}{51.1} & \multicolumn{1}{l|}{\textbf{82.7}} &  &  &                      &  &  &  &  &  \\ \cline{1-6}
\multicolumn{1}{|l|}{Avg.}  & \multicolumn{1}{l|}{54.4}     & \multicolumn{1}{l|}{52.6} & \multicolumn{1}{l|}{57.6} & \multicolumn{1}{l|}{52.6} & \multicolumn{1}{l|}{\textbf{91.4}} &  &  &                      &  &  &  &  &  \\ \cline{1-6}
\end{tabular}
\caption{\label{Tab:Class}  Results of the experiment (i), accuracy percentage on test data for the English and Portuguese corpora}
\end{table}

\textit{How much each model rely on the occurrence of non-logical words?}

With the full intersection of the vocabulary, experiment (ii), we have observed that the average accuracy improvement differs from model to model: Baseline, GRU, BERT$_{eng}$, LSTM and RNN present an average improvement of $17.6\%$, $9.6\%$, $5.3\%$, $4.25\%$, $1.3\%$, respectively. This may indicate that the recurrent models are relying more on noun phrases than BERT. However, since the difference is not significant, more investigation is required.

\textit{Can cross-lingual transfer learning be successfully used for the Portuguese realization of those tasks?}

As expected, when we fine-tuned BERT$_{multi}$ to the Portuguese version of the dataset we have observed an overall improvement. Most notably, in Tasks 6 and 7 we have achieved a new accuracy of $87.4\%$ and $92.3\%$ respectively.  Surprisingly, BERT$_{chi}$ is able to solve some simple tasks, namely Tasks 1, 2 and 4. But when trained on the mixed version of the dataset, Task 7, this pre-trained model had repeatedly present a random performance. 

\begin{figure}[h!]
    \centering
    \includegraphics[width=0.49\textwidth]{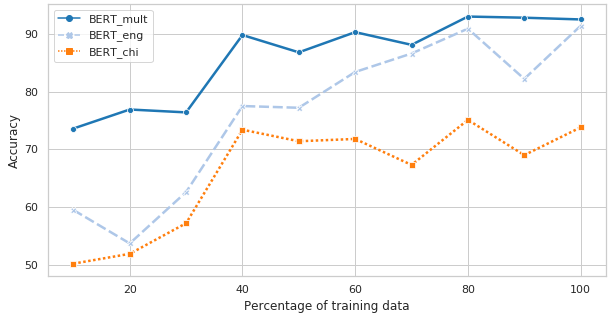}
  \caption{\label{Fig:acc3} Results of the experiment (iii), different pre-trained BERT versions tested on Portuguese corpus}
\end{figure}

One of the most important features observed by evaluating the different pre-training models is that although BERT$_{eng}$ and BERT$_{mult}$ show a similar result on the Portuguese corpus, BERT$_{eng}$ needs more data to improve its performance, as seen in Figure \ref{Fig:acc3}.

\textit{Is the dataset biased? Are the models learning some unexpected text pattern?}

\begin{figure}[h!]
    \centering
    \includegraphics[width=0.49\textwidth]{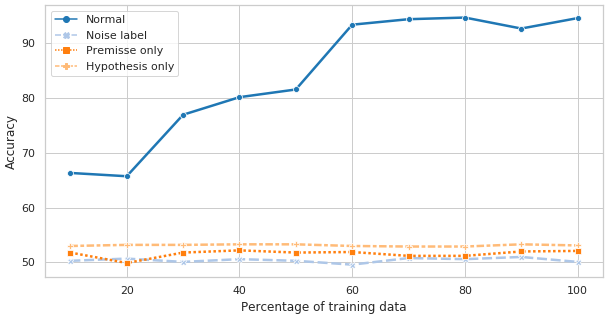}
  \caption{\label{Fig:acc4} Results of the experiment (iv), BERT$_{eng}$'s accuracy on the different versions of the data (English corpus)}
\end{figure}

By taking BERT$_{eng}$ as the best classifier, we repeated the training using all the listed data modification techniques. The results, as shown in Figure \ref{Fig:acc4}, indicate that BERT$_{eng}$ is not memorizing random textual patterns, neither excessively relying on information that appears only in the premise $P$ or the hypothesis $H$. When we applied it on these versions of the data, BERT$_{eng}$ behaves as a random classifier. 

\section{Discussion}

The results presented above are similar to the ones reported in \newcite{Goldberg19} : \textit{Transformer-based models like BERT can successfully capture syntactic regularities and logical patterns}.

These findings do not contradict the results reported on \newcite{Evans18,Tran18}, because in both papers, the Transformer models are trained from scratch, while here we have used models that were pre-trained on large datasets with the language model objective.

The results presented both in Table \ref{Tab:Class} and Figure \ref{Fig:acc3} seem to confirm our initial hypothesis on the effectiveness of transfer learning in a cross-lingual fashion. What has surprised us was the excellent results regarding Tasks 1, 2 and 4 when transferring structural knowledge from Chinese to Portuguese. We offer the following explanation for these results. Take the contradiction pair defined in the template language:

\begin{enumerate}
    \item[] $P:= \{ x_1 = \iota y \forall x_2 V(y, x_2) , V(x_1, x_3)\}$ (``$x_1$ \textit{is the person that has visited everybody, $x_1$ has visited $x_3$}")
    \item[] $H:= \lnot V(x_1, x_4)$ (``\textit{$x_1$ didn't visit $x_4$}")
\end{enumerate}

If we take one possible Portuguese realization of the pair above and apply the different tokenizers we have the following strings:  

\begin{enumerate}
    \item Original sentence: ``\textit{[CLS] \textbf{gabrielle} é a pessoa que \underline{visitou} \textbf{todo} mundo \textbf{gabrielle} \underline{visitou} luís [SEP] \textbf{gabrielle não} \underline{visitou} \textbf{ianesis} [SEP]}". 
    \item Multilingual tokenizer: ``\textit{[CLS] \textbf{gabrielle} a pessoa que \underline{visito $\#\#$u} \textbf{todo} mundo \textbf{gabrielle} \underline{visito $\#\#$u} lu $\#\#$s [SEP] \textbf{gabrielle no} \underline{visito $\#\#$u} \textbf{ian $\#\#$esis} [SEP]}"
    \item English tokenizer: ``\textit{[CLS] \textbf{gabrielle} a pe $\#\#$sso $\#\#$a que \underline{visit $\#\#$ou} \textbf{tod $\#\#$o} mundo \textbf{gabrielle} \underline{visit $\#\#$ou} lu $\#\#$s [SEP] \textbf{gabrielle no} \underline{visit $\#\#$ou} \textbf{ian $\#\#$esis} [SEP]}"
    \item Chinese tokenizer: ``\textit{[CLS] \textbf{ga $\#\#$b $\#\#$rie $\#\#$lle} a pe $\#\#$ss $\#\#$oa q $\#\#$ue \underline{vi $\#\#$sit $\#\#$ou} \textbf{to $\#\#$do} mu $\#\#$nd $\#\#$o \textbf{ga $\#\#$b $\#\#$rie $\#\#$lle} \underline{vi $\#\#$sit $\#\#$ou} lu $\#\#$s  [SEP] \textbf{ga $\#\#$b $\#\#$rie $\#\#$lle no} \underline{vi $\#\#$sit $\#\#$ou} \textbf{ian $\#\#$es $\#\#$is} [SEP]}"
\end{enumerate}

Although the Portuguese words are destroyed by the tokenizers, the model is still able to learn in the fine-tuning phase the \textit{simple} structural pattern between the tokens highlighted above. This may explain why the counting task (Task 4) presents the highest difficulty for BERT. There is some structural grounding for finding contradictions in counting expressions, but to detect contradiction in all cases one must fully grasp the \textit{meaning} of the multiple counting operators.

\section{Conclusion}

With the possibility of using pre-trained models we can successfully craft small datasets ($\sim$ 10K sentences) to perform fine grained analysis on machine learning models. In this paper, we have presented a new dataset that is able to isolate a few competence issues regarding structural inference. It also allows us to bring to the surface some interesting comparisons between recurrent neural networks and pre-trained Transform-based models. As our results show, \textit{compared to the recurrent models,  BERT presents a considerable advantage in learning structural inference. The same result appears even when fine-tuned one version of the model that was not pre-trained on the target language}.   

By the stratified nature of our dataset, we can pinpoint BERT's inference difficulties: \textit{there is space for improving the model's counting understanding}. Hence, we can either craft a more realistic NLI dataset centered on the notion of counting or modify BERT's training to achieve better results in the counting task. 

The results on cross-lingual transfer learning are stimulating. One possible area for future research is to check if the same results can be attainable using simple structural inferences that occur within complexes sentences. This can be done by carefully selecting sentence pairs in a cross-lingual NLI corpus like \newcite{xnli}. We plan to explore these paths in the future.

\end{document}